\documentclass[a4paper, 10pt, conference]{ieeeconf}      % Use this line for a4 paper

\IEEEoverridecommandlockouts                              % This command is only needed if 
\usepackage[english,french]{babel}
\usepackage[utf8]{inputenc}
\usepackage[T1]{fontenc}
\usepackage{url}
\usepackage{enumerate}

\usepackage{graphicx} % for pdf, bitmapped graphics files
\usepackage{amsmath} % assumes amsmath package installed
\usepackage{amssymb}  % assumes amsmath package installed

\newcommand{\M}[1]{\mathbf{#1}} % Bold for matrix
\newcommand{\V}[1]{\mathbf{#1}} % Bold for vector
\renewcommand{\epsilon}{\varepsilon}
\newcommand{\norm}[1]{\bigl\|\,#1\,\bigr \|}

\title{\LARGE \bf
Towards 3D karst underwater scene reconstruction \\ from rotating sonar data
}

\author{Georgios Evangelos Margaritis$^{1}$, Lionel Lapierre$^{2}$, Simon Rohou$^{2}$,\\ Zhi Yan$^{1}$, Andreas {Nüchter}$^{1,3}$, François Goulette$^{1}$% <-this % stops a space
\thanks{$^{1}$U2IS, ENSTA, Institut Polytechnique de Paris, 91120 Palaiseau, France,
        contact${:}$ {\tt\small francois.goulette@ensta.fr}}%
\thanks{$^{2}$Lab-STICC, ENSTA, Institut Polytechnique de Paris, 29200 Brest, France,
        contact${:}$ {\tt\small lionel.lapierre@ensta.fr}}%
\thanks{$^{3}$Informatics XVII -- Robotics, Julius-Maximilians-Universität Würzburg, Germany
        contact${:}$ {\tt\small andreas@nuechti.de}}%
}

\begin{document}

\maketitle
\thispagestyle{empty}
\pagestyle{empty}

%%%%%%%%%%%%%%%%%%%%%%%%%%%%%%%%%%%%%%%%%%%%%%%%%%%%%%%%%%%%%%%%%%%%%%%%%%%%%%%%
\begin{abstract}
Karst aquifers provide critical freshwater resources but pose significant hazards due to their complex and poorly understood subsurface geometry. Mapping these environments is challenging because sonar data from underwater exploration is sparse and noisy, while navigation estimates suffer from drift limiting standard 3D reconstruction methods.
We present a pipeline for reconstructing underwater karst conduits from a sonar profiler. We combine a continuous-time SLAM approach to correct trajectory drift with a novel two-stage deep learning method for surface reconstruction, producing an immersive and navigable 3D mesh for hydrogeological analysis.
\end{abstract}

%%%%%%%%%%%%%%%%%%%%%%%%%%%%%%%%%%%%%%%%%%%%%%%%%%%%%%%%%%%%%%%%%%%%%%%%%%%%%%%%
\section{INTRODUCTION}
Karst landscapes are an important part of the global environment. They form when soluble bedrock is dissolved.
They provide essential freshwater resources for around a quarter of the world population~\cite{goldscheider2020global}. These systems are characterized by underground networks of caves and conduits that form vast aquifers. These aquifers supply major cities and regions that are often affected by drought~\cite{ford2007karst}. However, the very properties that make them valuable also present significant hazards. Rapid, unfiltered flow of water can lead to sudden, catastrophic flash floods and quick dissemination of contaminants. In addition, the collapse of the underlying cavities poses a direct threat to infrastructure. The city of N\^{\i}mes, for example, faces dangerous floods directly linked to the dynamics of its underlying karst system, making the study of these environments a critical public safety priority~\cite{batiotguilhe2013contribution,marechal2008karst}.

Effective management and scientific understanding of these aquifers are fundamentally constrained by our limited knowledge of their subsurface geometry. This paper addresses the challenges of processing the data acquired by a robotic underwater vehicle equipped with a rotating sonar operated by a human diver.
Data gathered from sonar sensors during such missions is inherently sparse and noisy. Meanwhile, the trajectory calculated from onboard inertial and velocity sensors accumulates significant drift over long periods of exploration. These issues render standard 3D reconstruction techniques ineffective. Classical algorithms often fail to produce coherent surfaces from sparse data, and trajectory errors result in a distorted geometric representation of the environment.

We present a robust pipeline to transform this challenging raw sensor data into a high-fidelity, navigable 3D model of an underwater karst conduit. The contributions of this work are${:}$ (1) application of a continuous-time SLAM solution to correct severe trajectory drift (2) a novel, two-stage deep learning approach for this specific data
type and (3) a generation of an interactive 3D mesh suitable for virtual exploration by hydrogeologists.

\section{RELATED WORK}

Early autonomous mapping of inundated karst systems was pioneered by the DEPTHX AUV, whose array of single-beam sonars and inertial navigation produced the first geo-referenced 3-D model of Sistema Zacatón~\cite{Gary2008DEPTHX}. 

Subsequent acoustic SLAM efforts adopted smaller vehicles and mechanically  scanning imaging sonars${:}$ Mallios \textit{et al.} fused probabilistic scan-matching with DVL/IMU odometry to localize inside confined tunnels and incrementally build 2-D/3-D maps~\cite{Mallios2009PoseSLAM}. Because acoustic returns are sparse in narrow passages, vision-based methods were introduced, e.g., Weidner \textit{et al.} exploited the high-contrast contour of a diver’s lamp in stereo imagery to reconstruct hundreds of submerged conduit, using ORB-SLAM for pose estimation~\cite{Weidner2017Stereo}. Massone \textit{et al.} generalized this principle by calibrating a conical light source with a monocular camera, obtaining metrically-scaled meshes of fully flooded galleries~\cite{Massone2021Conical}.

\begin{figure*}
\centering
\includegraphics[width=0.43\textwidth]{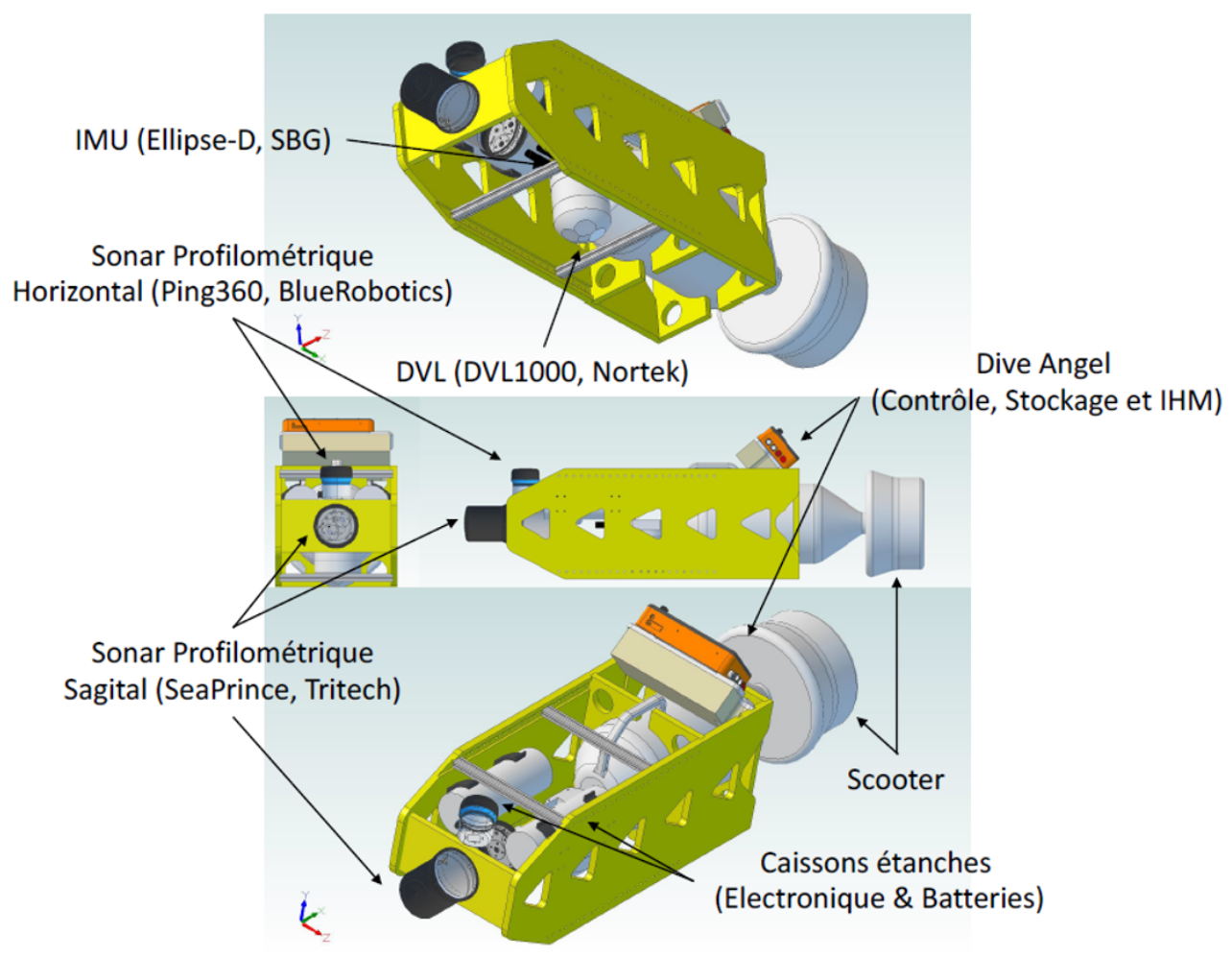}~
\includegraphics[width=0.55\textwidth]{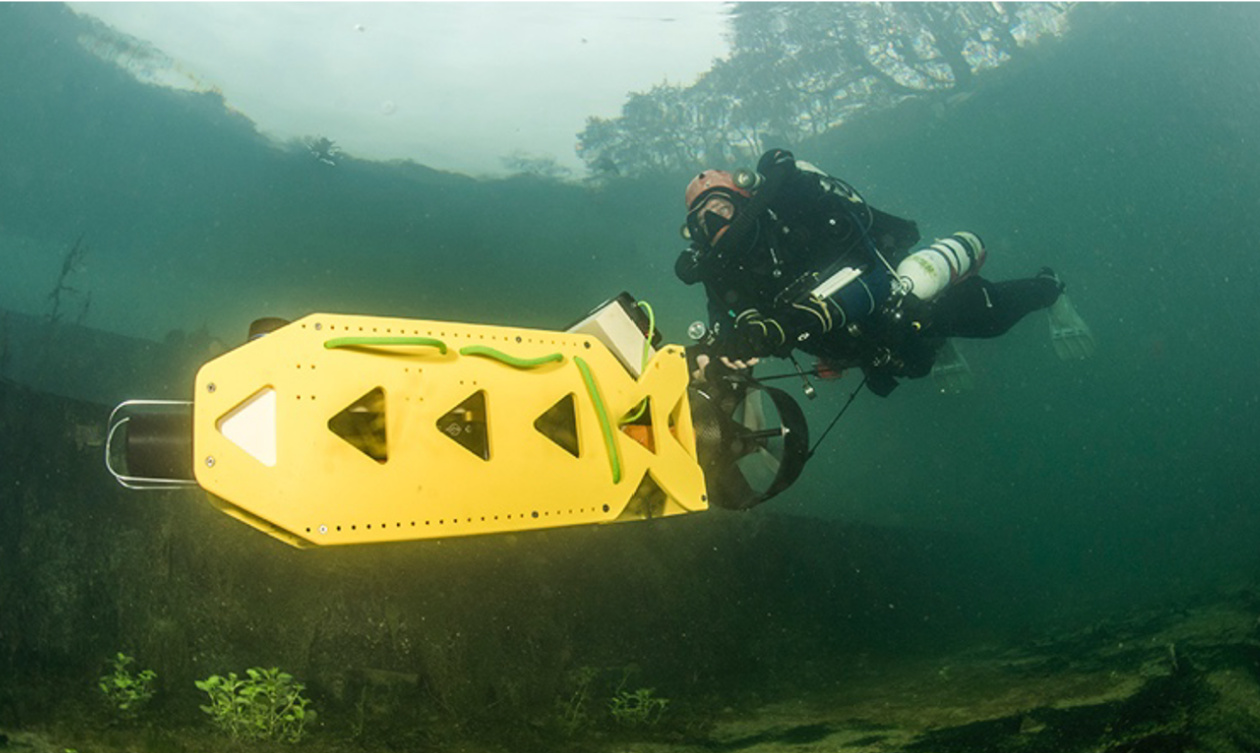}
\caption{Left: Overview of the robot system. Right: A professional diver operating the NavScoot2.}\label{fig:navscout}
\end{figure*}

Recent systems achieve robustness by tightly coupling multiple sensors. SVIn2 combines profiling sonar, stereo vision, IMU and depth sensing in a single factor-graph, delivering drift-free trajectories in wrecks and caves~\cite{Rahman2019SVIn2}; Wang \textit{et al.} augments that pipeline with an online multi-view stereo back-end, enabling real-time dense mesh generation on embedded CPUs~\cite{Wang2023Realtime}. On the acoustic side, Westman and Kaess resolved the elevation ambiguity of wide-aperture imaging sonar via a generative sensor model integrated in a TSDF framework~\cite{Westman2019WideAperture}, while McConnell and Englot used Bayesian object priors to predict missing depth in single-sonar returns, improving large-scale harbor reconstructions~\cite{McConnell2021Semantic}. Learning-driven perception has reached underwater subterranean environments${:}$ CaveSeg introduces a transformer-based network for semantic segmentation of cave imagery, supplying high-level cues (e.g., guide-lines, obstacles) that are fed back into SLAM and planning~\cite{Abdullah2023CaveSeg}. These studies show a clear evolution towards hybrid acoustic–visual SLAM systems augmented by learned priors, capable of delivering accurate 3-D reconstructions even in the most challenging karst environments.

\section{DATA ACQUISITION}

The Fontaine de N\^{\i}mes is one of the karst environments that has been explored. Studying the Fontaine de Nîmes karst system is crucial because it is the source of dangerous flash floods that directly threaten the city~\cite{batiotguilhe2013contribution,marechal2008karst}. These floods are severe in terms of volume and dynamics, leading to catastrophic consequences. For example, a devastating flood occurred on October 3, 1988, resulting in nine deaths and affecting 45,000 people. The estimated damage was \$600 million.

The data used in this paper have been acquired in March 2023 through the deployment of an underwater sensor system called NavScoot2 in the karstic network of the Fontaine de N\^{\i}mes. The NavScoot2 features several sensors, including two mechanically scanned imaging sonars (MSIS) for horizontal and vertical cave mapping, a Doppler velocity log (DVL), an inertial measurement unit (IMU), depth sensor, and a forward-looking echo-sounder with a second IMU integrated, cf. Fig.~\ref{fig:navscout}. Each sensor have been calibrated intrinsically and the CAD model has been used for deriving the extrinsic calibration.  The vertical profiling sonar is a Super SeaKing from Tritech with a scan resolution of 0.45$^\circ$. The IMU used is the standalone Ellipse A from SBG Systems, since it outperforms the one integrated in the echo-sounder.

\begin{figure*}
\centering
\includegraphics[height=0.32\linewidth,width=0.32\linewidth]{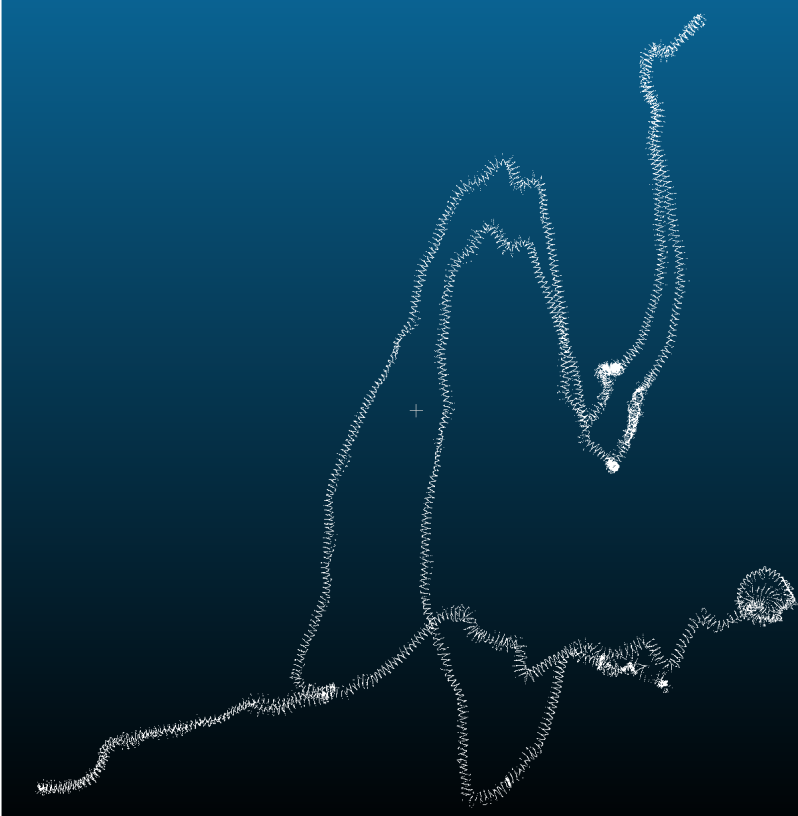}~
\includegraphics[height=0.32\linewidth,width=0.32\linewidth]{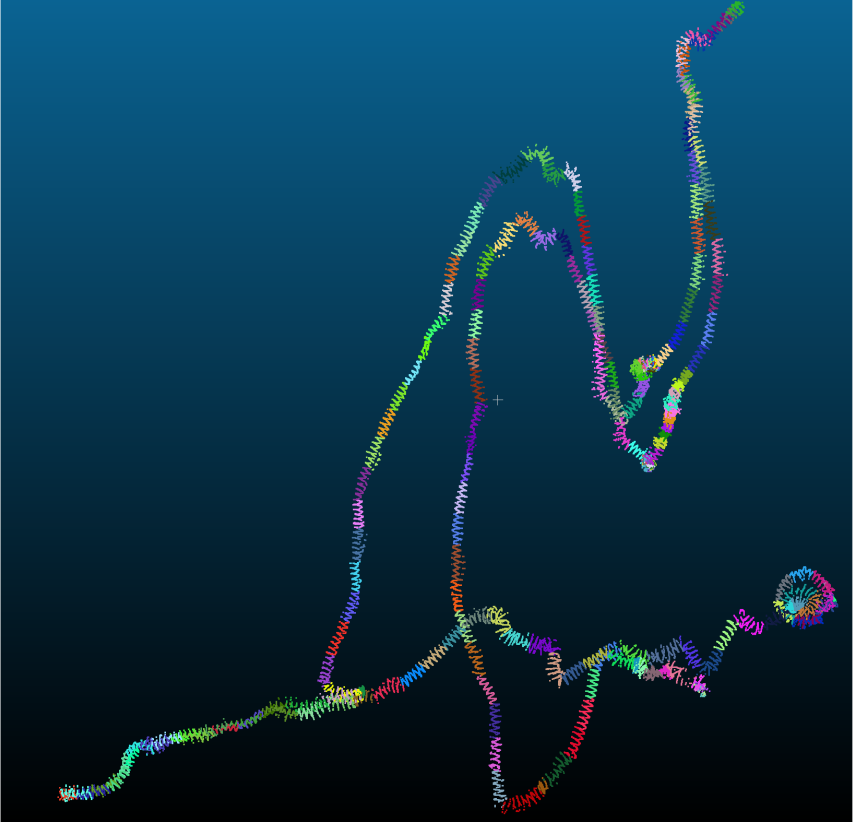}~
\includegraphics[height=0.32\linewidth,width=0.32\linewidth]{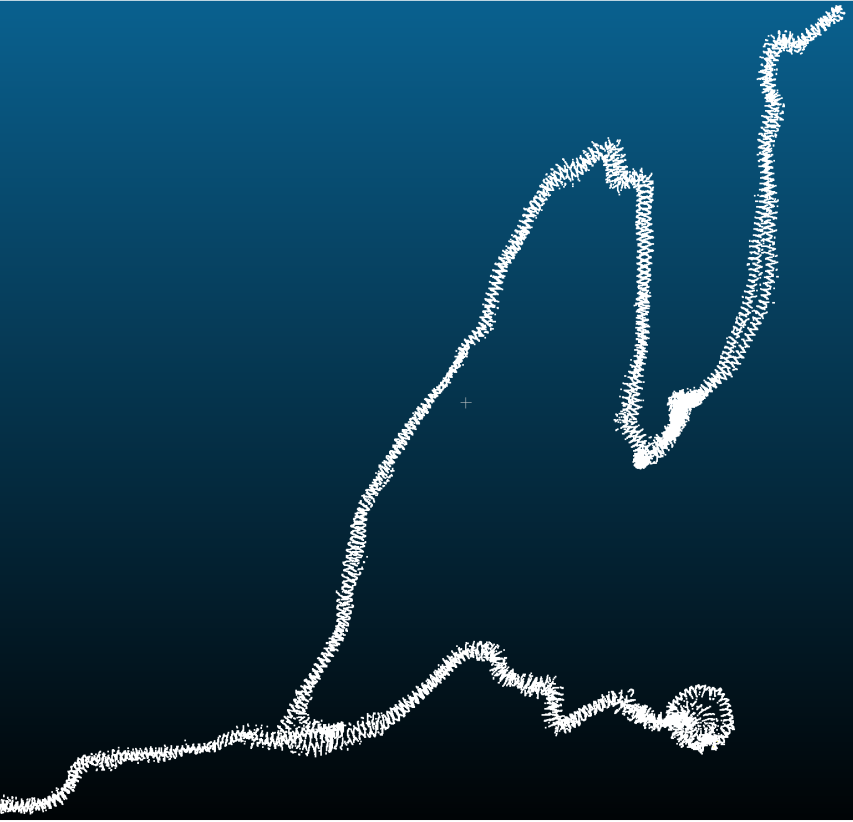}
\caption{Unwound sonar data as 3D point cloud in bird-eye view. Left: initial point cloud (cf. 3D view in Fig.~\ref{fig:sparse}), Middle: Split of the trajectory and the 3D point cloud into segments, Right: 3D point cloud after SLAM.}\label{fig:slam}
\vspace*{-2mm}
\end{figure*}

\section{3D RECONSTRUCTION}

The 3D reconstruction is performed in several steps: First, we join the distance measurements from the rotating sonar with trajectory information and process the data using a 6DoF SLAM algorithm for 3D point clouds. Then, we apply deep-learning based meshing algorithms. In the final step, the mesh is imported to the software Blender for final optimization and visualization.

\subsection{6DoF SLAM}

In an initial step, we unwind the data from the spinning vertical profiling sonar using the trajectory given by IMU and velocity data. Our unwinding method identifies the first significant return signal in each ping that surpasses a predefined intensity threshold. The focus on only the first return is critical${;}$ this signal corresponds to the shortest acoustic path and therefore represents the true,
direct distance to the nearest object. Any subsequent returns from the same ping are disregarded as they are likely multi-path echoes. The computed trajectory exhibits significant drift.
The resulting 3D point cloud consists of approx. 50,000 time-stamped 3D points, each observed from an individual pose. Fig.~\ref{fig:slam} (left) shows the obtained 3D point cloud. Please note the sparseness of the data and the deviation due to the drift.

\begin{figure}
\centering
\includegraphics[width=0.85\linewidth]{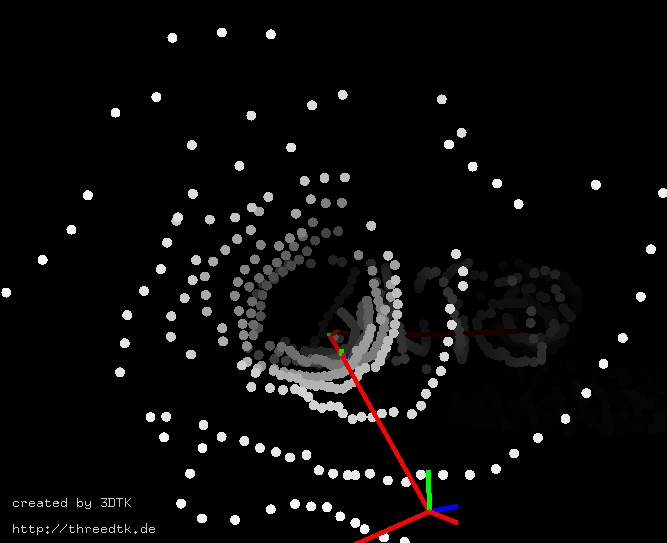}
\caption{The spinning profiler creates a spiral pattern inside the karst.}\label{fig:sparse}
\vspace*{-4mm}
\end{figure}

We employ the Continuous-time Simultaneous Localization and Mapping (SLAM) algorithm, namely 6D SLAM, available in 3DTK -- The 3D Toolkit~\cite{3DTK}. 6D SLAM works similarly to the well-known iterative closest points (ICP) algorithm, which minimizes the following error function
\begin{align}
E(\M R, \V t) & = \frac{1}{N} \sum_{i=1}^N \norm{\V m_i - (\M R \V d_i
 + \V t)}^2\label{eq:DMin}
\end{align}
to iteratively solve for an optimal rotation $\M T = (\M R, \V t)$, where the tuples $(\V m_{i}, \V d_i)$ of corresponding model $\V M$ and data points $\V D$ are given by minimal distance, i.e., $\V m_i$ is the closest point to $\V d_i$ within a close limit~\cite{Besl_1992}. Instead of the two-scan-Eq.~\eqref{eq:DMin}, we look at the $n$-scan case
\begin{align}
E =& \sum_{j \rightarrow k} \sum_i \left| \M{R}_j \V{m}_i + \V{t}_j - (\M{R}_k
\V{d}_i + \V{t}_k) \right|^2 ,\label{eq:LUM}
\end{align}
where $j$ and $k$ refer to scans of the SLAM graph, i.e., to the graph modeling the pose constraints in SLAM or bundle adjustment. If they overlap, i.e., closest points are available, then the point pairs for the link are included in the minimization. We solve for all poses at the same time and iterate like in the original ICP. Please note, while there are four closed-form solutions for the original ICP Eq.~\eqref{eq:DMin}, linearization of the rotation in Eq.~\eqref{eq:LUM} seems to be always required~\cite{CVIU2010}.

However, we do not have separate 3D point clouds, thus the time is coarsely discretized. This results in a partition of the trajectory into sub-scans or sub 3D point clouds that are treated rigidly.  
Thus, we first split the trajectory into sections, and match these sections using the 6D SLAM core. Here the SLAM graph is estimated using a heuristic that measures the
overlap of sections using the number of closest point pairs. After applying globally consistent scan matching on the sections the actual continuous-time or semi-rigid matching as described in~\cite{REMSEN2013} is applied, using the results of the rigid optimization as starting values to compute the numerical minimum of the underlying least square problem.

To ensure it’s findings meaningful, ``loop closures'', and not just matching points that are right next to each other, we use a time threshold. We only consider a pair of points as a match if they were recorded a certain amount of time apart and are closest points. This forces the algorithm to find correspondences between the whole point cloud. 
The trajectory was segmented into 100 overlapping patches. These meta-scans are visualized in Fig.~\ref{fig:slam} (middle). The application resulted in a significantly more precise, corrected trajectory of the robot's path that aligns closely with the ground truth, see Fig.~\ref{fig:slam} (right).
The search is achieved with a $k$-d tree, sped-up using OpenMP and makes use of the sparse Cholesky decomposition by~\cite{Davis_2005}. 

\subsection{Mesh Reconstruction}

Mesh reconstruction is the fundamental process of converting an unstructured point cloud into a structured and continuous surface representation, typically a polygonal mesh. A successful reconstruction enables essential downstream tasks such as realistic visualization, physical simulation and geometric analysis.

Well-known classical approaches to reconstruction like the Ball Pivoting Algorithm (BPA), Poisson Surface Reconstruction, and Delaunay triangulation do not produce reasonable meshes -- the 3D point cloud is too sparse. The BPA produces highly fragmented surfaces with significant holes, while Poisson Surface Reconstruction creates overly smoothed, topologically incorrect meshes that seal the conduit’s openings. The Delaunay/Voronoi-based method is unable to distinguish the thin conduit surface from its convex hull. This motivates our investigation into learning-based models capable of handling such challenging data. Many learning-based methods are designed to be robust to sparse, noisy, and incomplete point clouds, a scenario where traditional methods struggle~\cite{Farshian2023}. 

Our reconstruction pipeline is as follows${:}$ We use Neural Poisson Surface Reconstruction to produce an uniformly down-sampled version of the input point cloud. Afterwards, we use the learning-based Point Convolution for Surface Reconstruction for creating the mesh.

\subsubsection{Neural Poisson Surface Reconstruction (nPSR)}

Neural Poisson Surface Reconstruction (nPSR) introduces a novel deep learning architecture for reconstructing 3D shapes from oriented point clouds. The method’s primary innovation is the use of a Fourier Neural Operator (FNO) to solve the underlying Poisson equation, which allows it to be trained efficiently on low-resolution data and still perform effectively at high resolutions. This approach proves particularly robust in low-sampling scenarios, where traditional methods
often fail, while remaining competitive in high-sampling regimes~\cite{huang2023nksr}.

The authors build their method on established mathematical and machine learning concepts. The process begins with an oriented point cloud and their corresponding surface normal vectors. Their goal is to reconstruct/learn an indicator function, $\chi$, such that
\begin{align}
\nabla \chi = V(x)   
\end{align}
where the object’s shape is defined by assigning a value of 1 to points inside the object and 0 to points outside. I.e., it finds a scalar function $\chi$ whose gradient best approximates a vector field $V$. By taking the divergence of this relationship, the authors arrive at the Poisson equation${:}$
\begin{align}
\Delta \chi(x) = \nabla \cdot \nabla \chi (x) = \nabla \cdot V(x)
\end{align}
Solving this equation for $\chi$ allows for the reconstruction of the object’s shape. This is treated as a boundary value problem, where the function $\chi$ is assumed to be zero at the edges of the defined space.

To solve the Poisson equation within a deep learning framework, the authors employ a Fourier Neural Operator (FNO)~\cite{li2021fourier}. An FNO is a type of neural network specifically designed to learn solutions to partial differential equations
(PDEs). It works by learning a mapping between two function spaces. The architecture is composed of iterative \emph{Fourier blocks} that operate in the frequency domain. A key operation within these blocks is applying a learned weight tensor, $R$, to the Fourier-transformed input, which is analogous to a convolution but performed in Fourier space~\cite{li2021fourier}.

The continuous functions handled by the FNO must be discretized onto a grid. The authors highlight a critical advantage of FNOs${:}$ they are discretization-invariant. Because the network’s parameters are learned directly in Fourier space, they are not tied to a specific grid size. This property is what makes the nPSR model resolution-agnostic. It allows a model trained on low resolution data to be effectively transferred and evaluated on higher-resolution inputs without retraining, which is a significant benefit when dealing with memory-intensive 3D data. For their implementation, the authors use the Fast Fourier Transform (FFT) and truncate higher frequencies, keeping only a set number of Fourier modes~\cite{huang2023nksr}.

The complete pipeline takes an oriented point cloud as input and produces a final 3D mesh and a uniformly down-sampled point cloud using voxel grid down-sampling. The full method follows a three-stage process~\cite{huang2023nksr}${:}$
\begin{itemize}
    \item   
    Pre-Processing${:}$ Since the FNO architecture requires input on a uniform grid, the raw point cloud must be converted. This is achieved through point rasterization of the divergence field. The process first rasterizes the point locations into a voxel grid, applies a Gaussian smoothing filter, and then uses the normal vectors to compute the divergence field via finite differences.
    \item 
    Fourier Neural Architecture${:}$ The resulting voxel grid representing the divergence field is fed into the core of the model, which is an FNO composed of four 3D Fourier blocks. The network uses GELU non-linearities and dropout layers for regularization.
    \item 
    Post-Processing${:}$ The network outputs a reconstructed voxel grid. This output is refined using Otsu’s method to compute an optimal threshold value. Finally, the marching cubes algorithm is applied to the thresholded grid to extract the final watertight surface mesh~\cite{andrade-loarca2023neural}.
\end{itemize}

\begin{figure}
\centering
\includegraphics[width=0.605\linewidth]{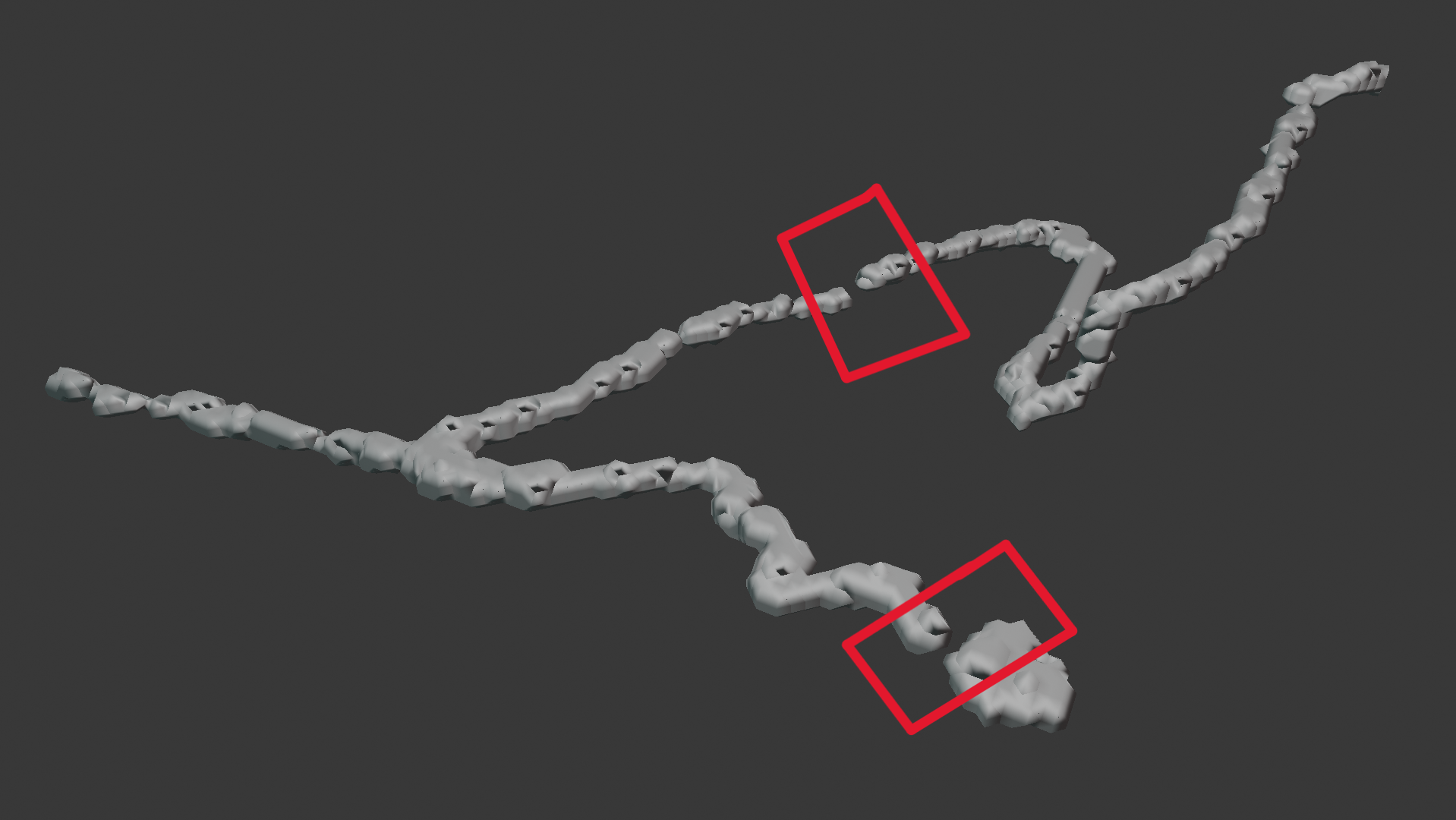}~
\includegraphics[width=0.36\linewidth]{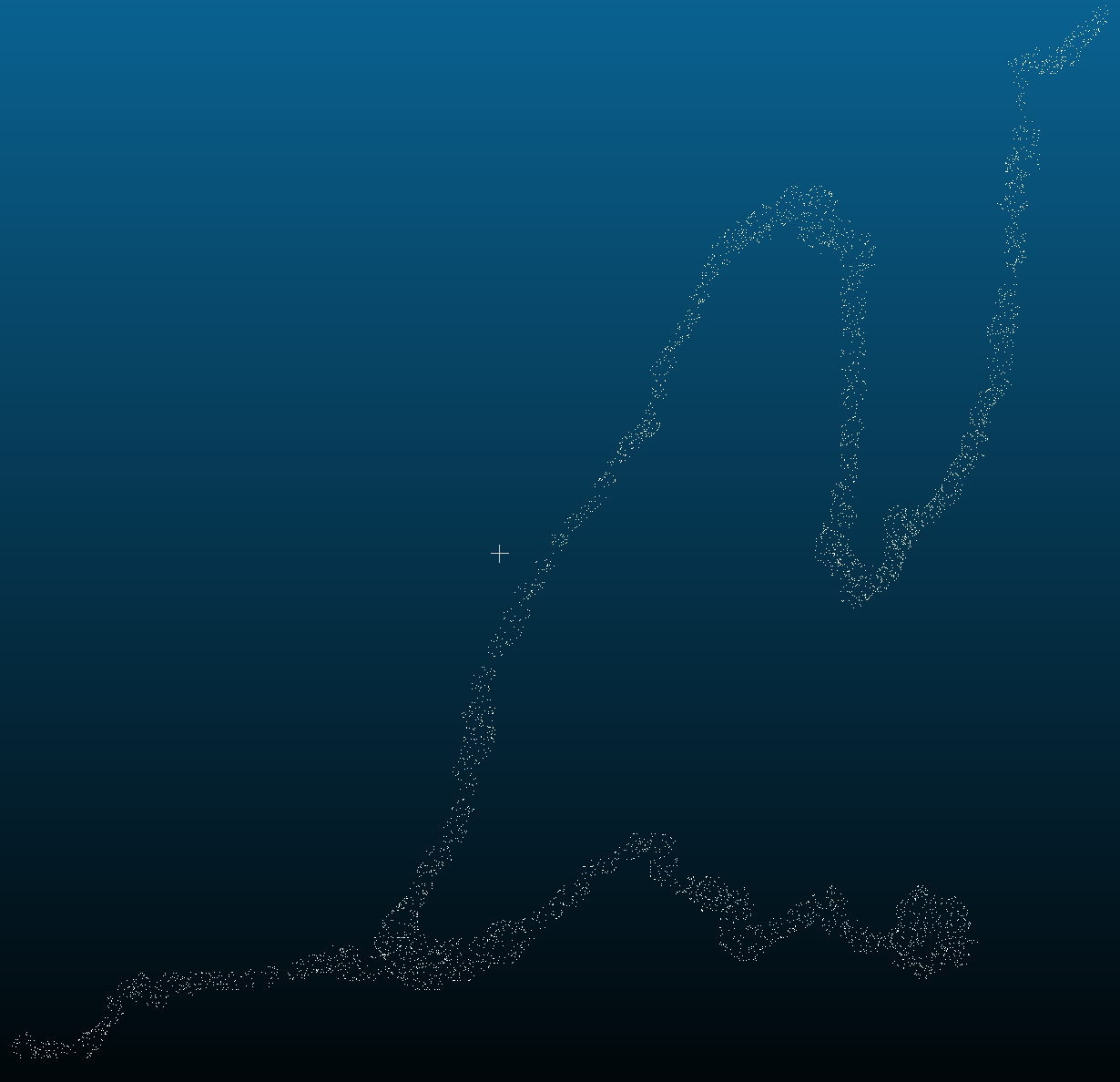}
\caption{Left: The 3D surface mesh reconstructed by the nPSR model. The result exhibits significant gaps and discontinuities (by red boxes). Right: The uniformly down-sampled point cloud generated by the nPSR pipeline.}\label{fig:npsr}
\end{figure}

We utilized the pre-trained version provided by the authors in the official repository. For inference, we used an oriented point cloud that was prepared with the NKSR model~\cite{huang2023nksr}. Upon running the nPSR inference pipeline with our data, the model generated two primary outputs${:}$ a reconstructed 3D surface mesh (see Fig.~\ref{fig:npsr}, left) and a uniformly down-sampled version of the input point cloud (see Fig.~\ref{fig:npsr}, right). The reconstructed mesh indicates a partial success; the model was able to approximate the main geometry of the conduit. Despite this, the result falls short of an optimal reconstruction, as it fails to resolve finer geometric details and is characterized by a lack of surface continuity and the presence of significant holes.

\subsubsection{Point Convolution for Surface Reconstruction}

Point Convolution for Surface Reconstruction, short POCO, is a novel, learning-based method for reconstructing 3D surfaces from point cloud. POCO leverages point cloud convolutions to compute feature-rich latent vectors directly at each input point. This approach concentrates the learned information where it is most crucial near the object’s surface and allows for greater scalability and detail in the final reconstruction~\cite{Boulch2021POCO}.

The core of the POCO method is an architecture designed to efficiently encode local and global shape information and decode it to determine the occupancy of any point in space. The process consists of two main phases${:}$ shape encoding and local decoding. Unlike methods that operate on a uniform grid, POCO begins by computing a latent feature vector for every single point in the input cloud.

The input point cloud is processed by a point convolution backbone~\cite{Boulch2021POCO}. This network produces a 32-dimensional latent vector $\V z_{\V p}$ for each input point $\V p$, effectively encoding the local geometry surrounding that point within a larger receptive field. This step ensures that the learned features are directly associated with the sampled surface points.

To determine if an arbitrary query point $\V q$ in space is inside or outside the surface, i.e., its occupancy, POCO uses a learned, attentive interpolation scheme based on the features of nearby input points and executes the following four steps${:}$ Neighborhood Selection, Relative Encoding, Attentive Interpolation, and Occupancy Prediction:
\begin{enumerate}[(i)]
\item 
 For any given query point $\V q$, the method identifies its $k$ nearest neighbors from the original input point cloud (where $k$ = 64 in the experiments).
\item 
 The latent vector $\V z_{\V p}$ of each neighbor $\V p$ is concatenated with the relative coordinates ($\V q - \V p$). This combined vector is then passed through a small multilayer perceptron (MLP) to produce a relative latent vector $\V z_{\V p, \V q}$. This step makes the feature representation dependent on the query point’s position relative to the surface points. 
 \item 
 Instead of a simple averaging, POCO learns to weight the influence of each neighbor. An attention mechanism assigns a significance weight $s_{\V p, \V q}$ to
each relative latent vector $\V z_{\V p, \V q}$ . These weights are then used to compute a final interpolated feature vector $\V z_{\V q}$ for the query point as a weighted sum of the relative vectors in its neighborhood.
\item 
Finally, the interpolated vector $\V z_{\V q}$ is passed through a linear layer to produce occupancy logits, which are converted into a probability indicating
whether the point $\V q$ is inside the shape. The surface is then defined as the 0.5-probability iso-surface, which is extracted using an algorithm like Marching Cubes~\cite{lorensen1987marching}
\end{enumerate}
The POCO model was trained on several standard 3D shape and scene datasets to demonstrate its robustness and generalization capabilities.

We utilized the pre-trained models released with the POCO paper~\cite{Boulch2021POCO}. This allowed for a direct application of their state-of-the-art method to our specific use case of underwater karst conduits. During the inference process, the output resolution for the Marching Cubes algorithm was an adjustable parameter. 
We observed that while higher resolutions yielded a marginal improvement in the mesh quality, the difference from lower-resolution results was not substantial. The qualitative results from the reconstruction of our point clouds are as follows${:}$
\begin{figure*}
\centering
\includegraphics[width=0.495\linewidth]{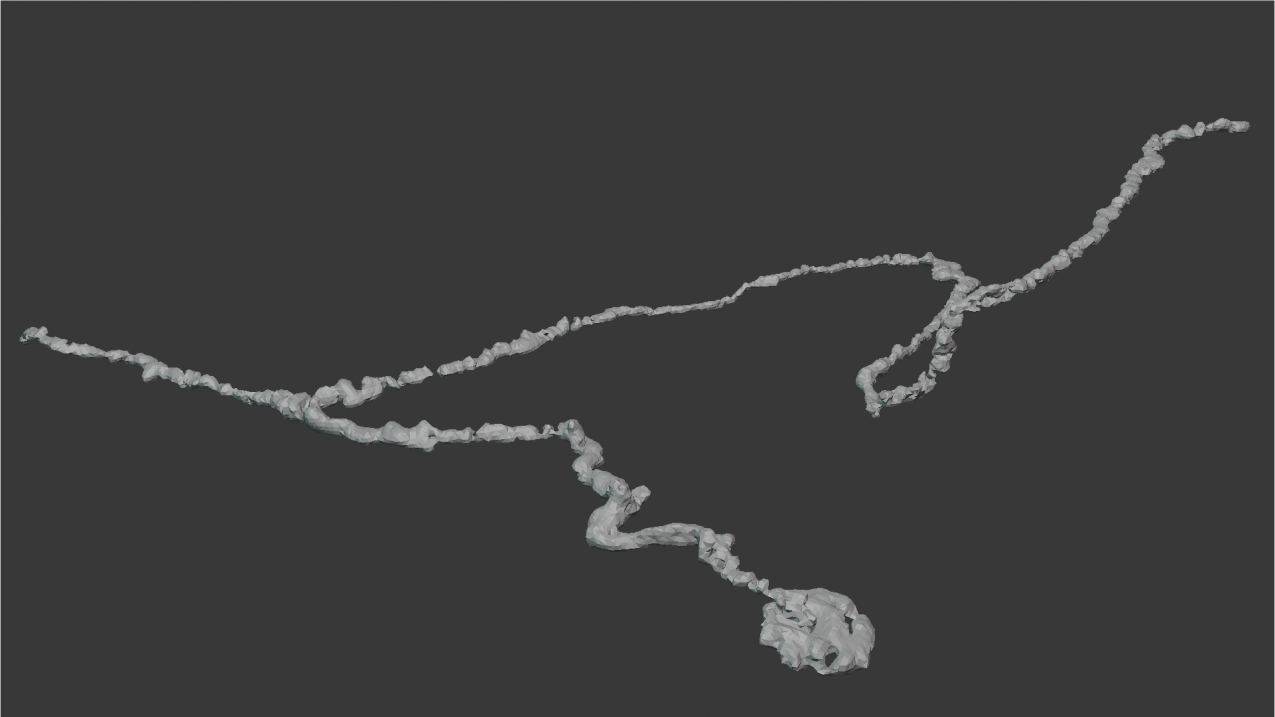}~
\includegraphics[width=0.495\linewidth]{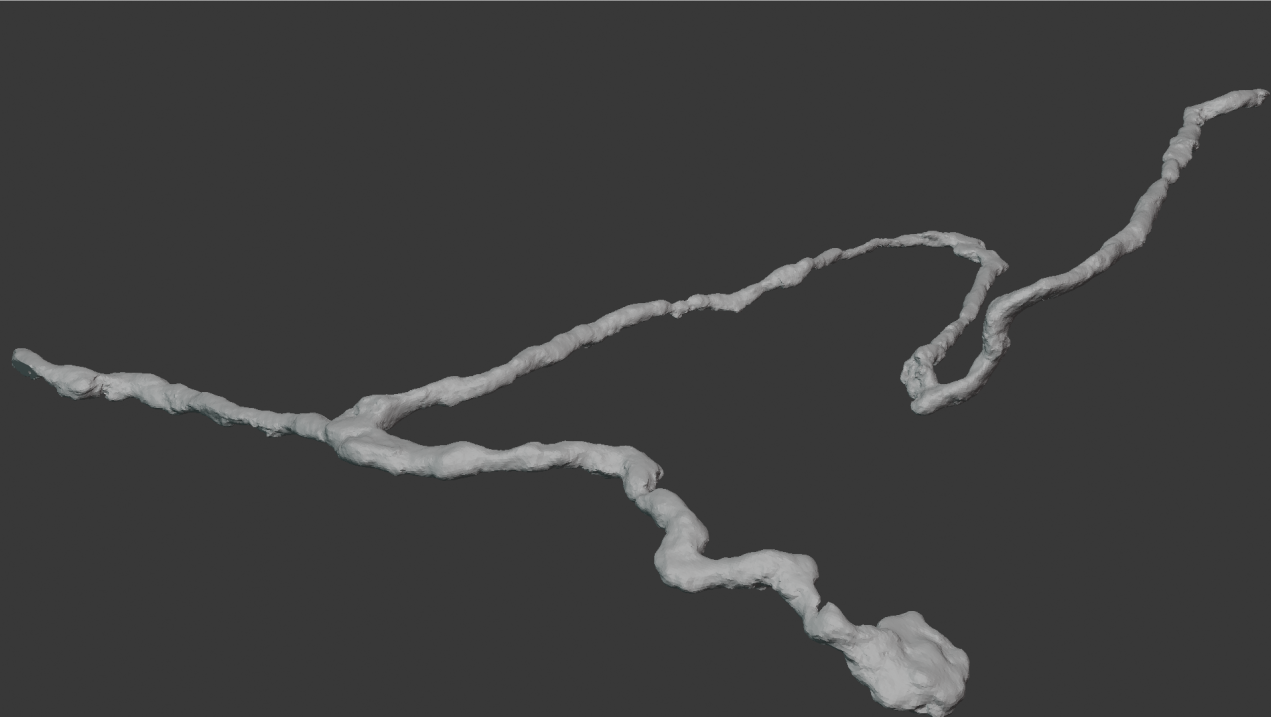}~
\caption{Qualitative comparison of surface reconstructions of the underwater karst conduit using pre-trained POCO models. The reconstruction from the down-sampled 3D point Neural Poisson Surface Reconstruction is considered our most accurate and reliable result (right), showing significant improvement in surface quality
and detail over the result from the original point cloud (left).}\label{fig:final_overview}
\end{figure*}
The reconstruction from the original 3D point cloud accurately describes the general geometry of the karst conduit. However, the resulting mesh exhibits some
small discontinuities and gaps in areas of sparse data (see Fig.~\ref{fig:final_overview}, left). 
The down-sampled 3D point neural Poisson surface reconstruction (nPSR) yields the best qualitative reconstruction. The resulting mesh describes the karst conduit with high fidelity, presenting a clean and detailed surface without noticeable artifacts or discontinuities. This result is considered our most accurate and reliable reconstruction (see Fig.~\ref{fig:final_overview}, right).

\subsection{Visualization}

Creating a continuous surface mesh and rendering it provides a tangible and intuitive model of the underwater environment, which is crucial for understanding the groundwater dynamics and preferential flow paths that define the karst aquifer. Furthermore, a realistic visualization enables ``virtual exploration'', allowing researchers to analyze the complex and otherwise inaccessible conduit network repeatedly without the physical risks and limitations faced by human divers. To
achieve this, we import the mesh into the 3D modeling software Blender. To ensure a continuous and solid surface suitable for interactive navigation, Blender’s Voxel Remesh tool was applied. This function completely reconstructs the mesh topology by first converting the input shape into a high-resolution 3D grid
of cubes, known as voxels. A new, manifold (watertight) mesh is then generated by creating a surface that effectively ``shrink-wraps'' this volumetric representation. This process not only creates a uniform and evenly distributed set of polygons, resulting in a clean, high-detailed, and solid model for visualization.
\begin{figure*}
\centering
\includegraphics[width=0.32\linewidth]{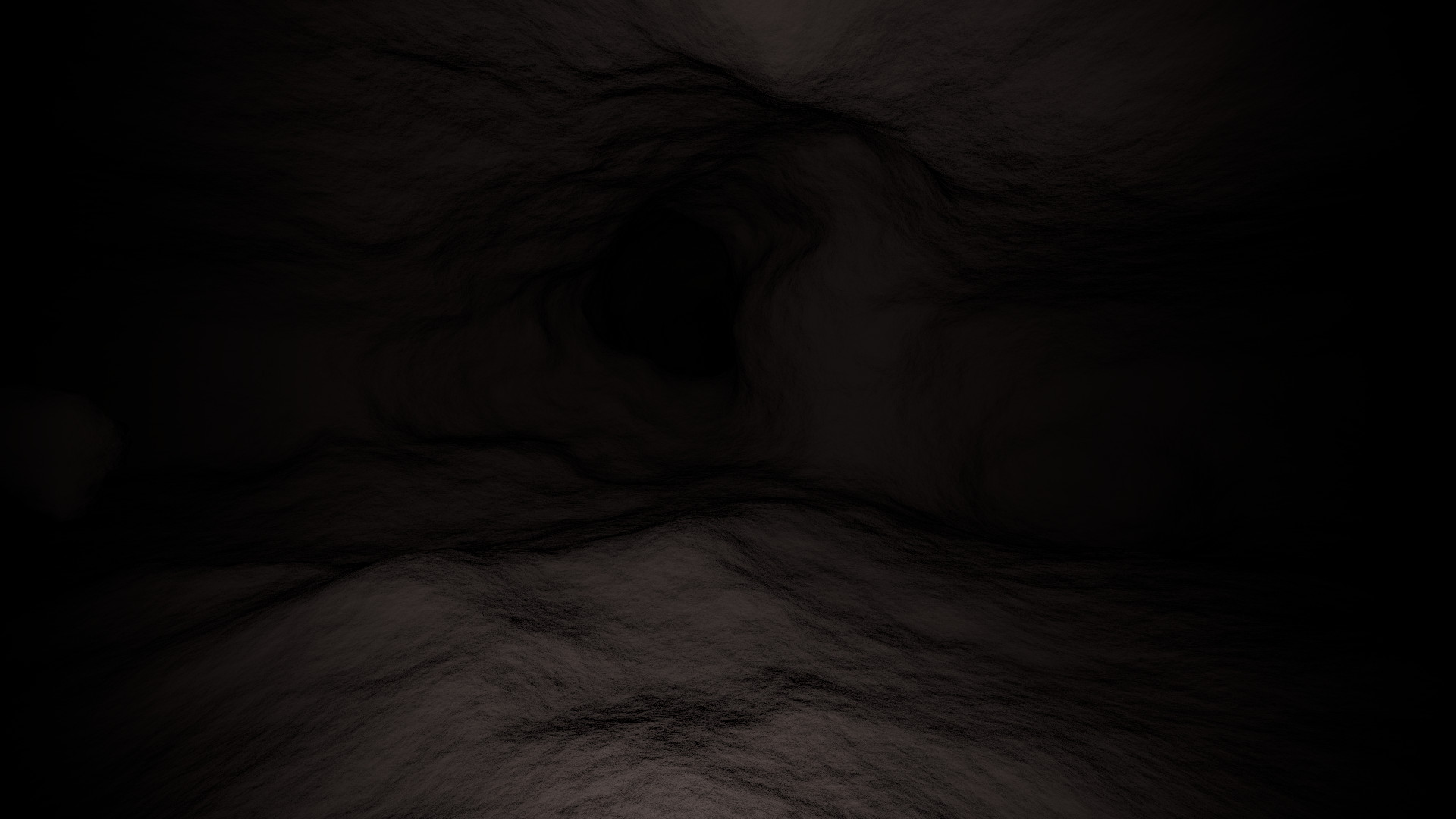}~
\includegraphics[width=0.32\linewidth]{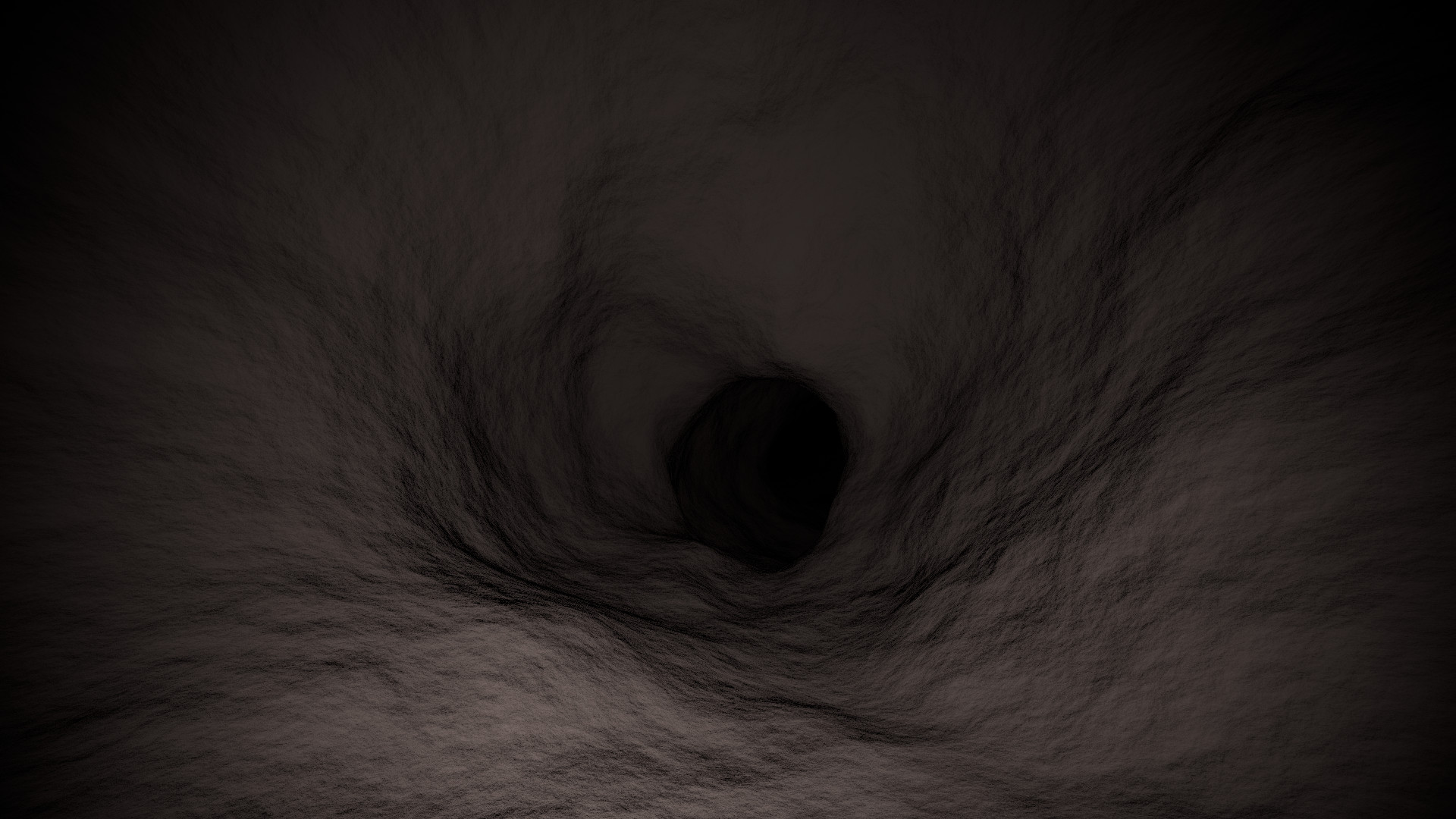}~
\includegraphics[width=0.32\linewidth]{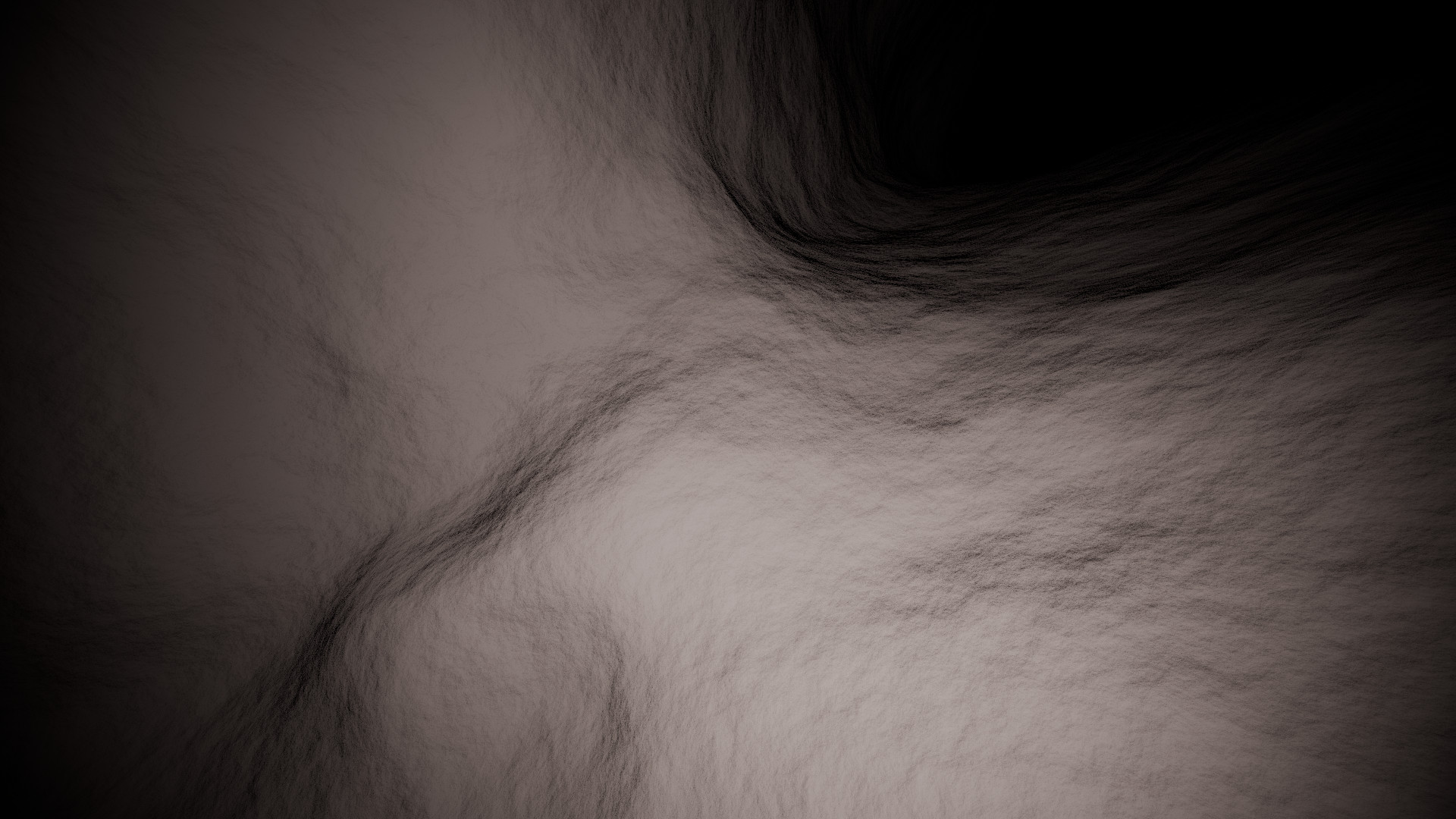}
\caption{A series of first-person views from the interactive visualization created in Blender. The images demonstrate the capability for virtual exploration and the implementation of dynamic lighting to enhance spatial awareness within the reconstructed karst conduit. Left: The initial first-person starting view at the entrance of the virtual conduit. Middle: A view from a different position within the model, looking along a section of the conduit. Right: A downward view illustrating the proximity based lighting effect, where surfaces closer to the camera are illuminated more strongly.
  \texttt{https://youtu.be/KduRSlPYo6Q}
}\label{fig:blender}
\end{figure*}
This mesh is then used to create an interactive, first-person visualization of the conduit's interior, enabling virtual exploration through ``fly'' navigation and a more immersive ``walk with gravity'' mode (see Fig.~\ref{fig:blender}). To simulate the experience of a diver exploring the dark underwater environment, the conduit's interior has been made dark and a single light source was attached to the camera's point of view, mimicking the diver’s point of view.

\section{CONCLUSIONS}

This paper has presented a complete pipeline for reconstructing underwater karst conduits from sparse and noisy sonar data. By integrating rotating sonar measurements with trajectory estimates and refining them through a 6DoF SLAM framework, we mitigate drift and obtain a consistent 3D point cloud. A deep learning–based meshing approach is then applied to generate a continuous surface, followed by final optimization and visualization in Blender.

The resulting meshes demonstrate that reliable geometric representations can be recovered despite challenging acquisition conditions, enabling improved visualization and analysis of karst systems. This approach provides a practical tool for hydrogeological exploration and lays the groundwork for future developments in autonomous mapping and higher-fidelity reconstruction of complex subsurface environments.

Needless to say, a lot of work remains to be done. In future work we seek towards an autonomous solution, which would drop the need of a professional diver. Furthermore, an automatic underwater mapping system shall include recording and processing of images with a calibrated camera to enhance the immersive 3D reconstruction. 


%%%%%%%%%%%%%%%%%%%%%%%%%%%%%%%%%%%%%%%%%%%%%%%%%%%%%%%%%%%%%%%%%%%%%%%%%%%%%%%%

%%%%%%%%%%%%%%%%%%%%%%%%%%%%%%%%%%%%%%%%%%%%%%%%%%%%%%%%%%%%%%%%%%%%%%%%%%%%%%%%

%%%%%%%%%%%%%%%%%%%%%%%%%%%%%%%%%%%%%%%%%%%%%%%%%%%%%%%%%%%%%%%%%%%%%%%%%%%%%%%%

\section*{ACKNOWLEDGMENT}

The funding for the Hi!Paris International Visiting Chair at U2IS, ENSTA for Andreas Nüchter is gratefully acknowledged.

\bibliographystyle{ieeetr}
\bibliography{references,additional_bib,andreas_publications}

%%%%%%%%%%%%%%%%%%%%%%%%%%%%%%%%%%%%%%%%%%%%%%%%%%%%%%%%%%%%%%%%%%%%%%%%%%%%%%%%
\end{document}